\documentclass[journal]{IEEEtran}
\usepackage{xcolor,soul,framed} 
\colorlet{shadecolor}{yellow}
\usepackage[pdftex]{graphicx}
\graphicspath{{../pdf/}{../jpeg/}}
\DeclareGraphicsExtensions{.pdf,.jpeg,.png}
\usepackage[cmex10]{amsmath}
\usepackage{array}
\usepackage{mdwmath}
\usepackage{mdwtab}
\usepackage{eqparbox}
\usepackage{url}
\usepackage{multirow}
\usepackage{color}
\usepackage{subcaption}
\makeatletter

\newcommand{\Rmnum}[1]{\expandafter\@slowromancap\romannumeral #1@}
\makeatother

\begin{document}
\bstctlcite{IEEEexample:BSTcontrol}
\title{Human Recognition Using Face in Computed Tomography}

  \author{Jiuwen Zhu,
      Hu Han,~\IEEEmembership{Member,~IEEE,}
      and~S. Kevin Zhou,~\IEEEmembership{Fellow,~IEEE}

  }  


\maketitle
\begin{abstract}
\boldmath
With the mushrooming use of computed tomography (CT) images in clinical decision making, management of CT data becomes increasingly difficult. From the patient identification perspective, using the standard DICOM tag to track patient information is challenged by issues such as misspelling, lost file, site variation, etc. In this paper, we explore the feasibility of leveraging the faces in 3D CT images as biometric features. Specifically, we propose an automatic processing pipeline that first detects facial landmarks in 3D for ROI extraction and then generates aligned 2D depth images, which are used for automatic recognition. To boost the recognition performance, we employ transfer learning to reduce the data sparsity issue and to introduce a group sampling strategy to increase inter-class discrimination when training the recognition network. Our proposed method is capable of capturing underlying identity characteristics in medical images while reducing memory consumption. To test its effectiveness, we curate 600 3D CT images of 280 patients from multiple sources for performance evaluation. Experimental results demonstrate that our method achieves a 1:56 identification accuracy of 92.53\% and a 1:1 verification accuracy of 96.12\%, outperforming other competing approaches.
\end{abstract}


\IEEEpeerreviewmaketitle

\section{Introduction}

\IEEEPARstart{C}{omputed} tomography (CT) has become an essential imaging modality for medical imaging since its introduction into clinical practice in 1973~\cite{Description_of_system}. It can produce tomographic images combining many X-ray measurements taken from different angles and thus provide internal body structure and pathological characteristics. It has often been used for disease screening of various diseases in head, brain, lung heart, etc. The significant use of CT in clinical diagnosis also brings challenges in CT image management. The CT images may be lost partially or corrupted due to erroneous operations. For example, the CT images are usually stored in the form of Digital Imaging and Communications in Medicine (DICOM), which can contain not only the CT data, but also the users' name and ID. When there are input errors with the name or ID information, or these fields get corrupted, the correspondence between the multiple CT scans may no longer be reliable, leading to difficulties when the doctors need to review all the historical CT scans. In addition, people may go to more than one hospital in order to get a confirmed diagnosis result. Doctors need to confirm whether the CT images obtained in other hospitals belong to the same patient. There is an increasing demand for automatically associating different CT scans to the corresponding subjects, so that the 3D CT images are identified, registered, and archived. 

Exploring biometrics in medical images has already attracted some research interest~\cite{agrafioti2009medical, poree2016ECG}. Medical biometrics such as electrocardiogram (ECG), electroencephalogram (EEG), or blood pressure signals have been considered to be informative modalities for the next generation of biometric characteristics~\cite{agrafioti2009medical}.
In addition, exploration of biometric features in medical images is able to give assistance to the protection of medical data and provide references for improving biometric security system. For example, ECG has been employed in biometric applications~\cite{poree2016ECG}. A biometric authentication algorithm by utilizing ECG is proposed to unlock the mobile devices~\cite{ECG_mobile_devices}. To the best of our knowledge, our work is the first attempt to discover a biometric identifier from 3D CT images.

The challenge of our work is to capture the most subject discriminative features from 3D CT images.
CT images are usually stacked by a series of slices, which can be represented by a volume that is a collection of voxels. We note that the number of slices, slice spacing, and voxel resolution of different CT images are generally different, which can cause big diversities of CT images. Moreover, 3D CT images are susceptible to a number of artifacts, such as patient movement~\cite{bhowmik2012mitigating}, representation method~\cite{goldman2008principles} and radiation dose~\cite{barkan2013adaptive}. To achieve the best performance, we focus on the face area in CT, which occupies the majority of CT images, especially in head and neck CT. 
Furthermore, face recognition has been the prominent and increasingly mature biometric technique for identity authentication~\cite{wang2018deep}.
Thus, we utilize the 3D CT face as our biometrics.

We investigate existing 3D data processing methods. 
3D volumetric data has endowed a variety of representations and corresponding learning methods. For example, 3D convolutional neural networks (CNNs)~\cite{vnet,2015voxnet} are proposed to directly deal with volumetric data. In~\cite{vnet}, V-Net, a special type of CNN, is proposed for the 3D clinical prostate magnetic resonance imaging (MRI) data for specific segmentation tasks. However, such methods usually require unifying input data (i.e., a fixed number of voxels and volume size), which may not be suitable in dealing with diverse 3D CT images. Furthermore, due to the computational cost of 3D convolution and data sparsity, volume representation is limited by its resolution. Vote3D~\cite{wang2015voting} is a method proposed for dealing with sparsity problem. The main idea is utilizing a sliding window and a voting mechanism to handle 3D volume data. However, the operation is still based on sparse data, which may result in loss of detailed information. Again, these 3D CNN-based methods require unifying input, which may not be suitable for our irregular CT images. PointNet~\cite{qi2017pointnet} and PointNet++~\cite{qi2017pointnet++} are proposed for irregular 3D point cloud data representation learning, and have reported promising performance on classification and segmentation. However, such superiority is dispensable to the 3D data that can be regularly arranged. There is a size limitation of PointNet based method, i.e., usually smaller than 2048 points. Multiview CNNs~\cite{su2015multiView_cnn} is proposed to render 3D data into 2D images and then applying 2D CNNs for classification tasks. By their design, promising performance has been reported in~\cite{su2015multiView_cnn} in shape classification and retrieval tasks.

Since there are a number of excellent classification networks for 2D image-based classification and identification tasks, we also propose to perform 3D CT based person identification by projecting CT into multi-view 2D renderings.

Unlike common computer vision tasks like image classification and face recognition, which usually have millions of images, the typical amount of medical images is relatively small. To cope with this, one popular solution is to apply transfer learning~\cite{quantifyingOsteoarthritis, Evidenceschizophrenia}, which has been proved efficient in quantifying the severity of radiographic knee osteoarthritis~\cite{quantifyingOsteoarthritis} and improving schizophrenia (SZ) classification performance~\cite{Evidenceschizophrenia}. A popular and simple strategy is to fine-tune the pretrained network on the target dataset. In such cases, finding a suitable source domain dataset and minimizing the gap between source and target domain is crucial. In our tasks, we utilize the ample amount of face depth images obtain with RGB-D sensors to perform transfer learning towards CT based face recognition.

We summarize the main contributions of this work as follows:
\begin{enumerate}
\item To the best of our knowledge, this is the first attempt to explore the biometric characteristic of 3D CT images for human recognition, and we show 3D CT images do contain subject discriminative information, and achieve 92.53\% rank-1 identification rate on a dataset with 600 CT images.
\item We propose an automatic processing pipeline for 3D CTs, which first detects facial landmarks in 3D CTs for ROI extraction and then generates aligned 2D depth images projected from 3D CTs.
\item We employ transfer learning and a group sampling strategy to handle the small data issue, which is found to improve the person recognition performance by a large margin.
\item We validate the effectiveness and robustness of our method using datasets from multiple sources. The results show our model is able to capture the discriminative characteristics from CT images.
\end{enumerate}

The remainder of this paper is organized as follows. We briefly review related works in Section \Rmnum{2}. The proposed 3D CT based recognition method is detailed in Section \Rmnum{3}. In Section \Rmnum{4}, we provide the experiments on several public 3D CT datasets. Discussions of the method and results are provided in Section \Rmnum{5}. We finally conclude this work in Section \Rmnum{6}.

\section{Related work}

\subsection {Point Cloud Based 3D Modeling}

PointNet~\cite{qi2017pointnet} is proposed for data representation learning from point cloud, which is data format consisting of point coordinates listed irregularly. PointNet is a novel convolutional neural network that takes point cloud as input and learns the spatial pattern of each point, and finally aggregates the individual features into a global view. The network structure is simple but efficient. It has been successfully applied in object classification, partial segmentation, and scene semantic analysis tasks. A hierarchical feature learning method PointNet++~\cite{qi2017pointnet++} is proposed to better capture local structure and improve the ability of managing the variable density of point cloud. The main idea is to extract features to a higher dimension in a global pattern from a small scale. It introduces a strategy to divide and group the input points into some overlapping local regions by measuring the distance in a metric space and produces higher level features by utilizing PointNet. Such a method can achieve good performance in dealing with irregular point cloud data. However, such superiority is dispensable to the 3D data that can be regularly arranged. Moreover, it has a strong demand for high computational consumption and large memory.

VoxelNet~\cite{zhou2018voxelnet} proposed by Apple Inc offers a novelty for handling point cloud data. It encodes point cloud data to a descriptive volumetric representation by introducing a voxel feature encoding layer. Then the problem can be translated to the task of operating 3D volume data. But it is proved to perform well only in LiDAR dataset and its performance is uncertain for other tasks. PointGrid~\cite{le2018pointgrid} integrates point and grid for dense 3D data. It holds a constant number of points in a grid cell and gets a local, geometrical shape representation. However, the random screening of points causes a loss of detailed information, thus difficult to recognize the features on a small scale.

We note that those methods are usually applied in object identification tasks, in which the target objects have macroscopic differences in shapes or colors and do not request for high resolution of source data. It may not be suitable for those tasks that require finer granularity such as facial texture.

\subsection {2D Rendering Based 3D Modeling}
3D shape models can be naturally encoded by a 3D convolutional neural network. The method may be limited by computational complexity and restricted memory. In addition, 3D model is usually defined on its surface, which ignores the observed information reflected in the pixels of 2D images~\cite{qi2016volumetricAndMultiviewOn3dData}. Su et al. prove that applying a 2D CNN on multiple 2D views can achieve a better recognition result~\cite{Multi-view_convolutional_neural_networks_for_3d_shape_recognition}. They build classifiers of 3D shapes based on 2D rendered images, which outperform the classifiers directly based on 3D representations. Compared to the general 3D volume-based method, the 2D CNN on multiple 2D images consumes less memory. It allows an increased resolution and better expresses the fine-grained pattern. Another superiority of using 2D representations is that it offers an additional training data augmentation, since the 3D model of a subject is a single data point, which makes the 3D training dataset short of richness and multiformity. By projecting a 3D model into a 2D space, we can easily obtain a series of related 2D images for training, which is beneficial for network training. 2D images have wide sources and have already organized as massive image databases such as ImageNet~\cite{imagenet} which offers a plethora of information conducive to pre-training powerful features. The depth information and observed information can also be generated for improved performance~\cite{Multi-view_convolutional_neural_networks_for_3d_shape_recognition}. It has been also proved to achieve high results in recent retrieval benchmarks~\cite{largescale3dshaperetrieval}. Due to its advantages, it has designed for some classification and detection tasks. For example, Hegde et al.~\cite{fusionnet} propose FusionNet for 3D object classification after generating multiple representations. Deng et al.~\cite{largescale3dshaperetrieval} develop a 3D Shape retrieval method named CM-VGG by clock matching and using convolutional neural networks. Kalogerakis et al.~\cite{multi-viewFCNs}, inspired by the projecting strategy, develop a deep architecture for 3D objects segmentation tasks by combining multi-view FCNs and CRF. It learns to adaptively select part-based views to obtain special view-based shape representations.

2D CNN has achieved a series of breakthroughs in image classification and segmentation tasks. It can naturally integrate low/medium/high level features and classifiers in an end-to-end manner. The network is flexible due to different ways of stacking layers. Among many promising methods, ResNet~\cite{he2016resnet} proposed by He et.al is an efficient and powerful structure which can be easily implemented and modified. It is capable of solving the network degradation problem as the depth of the network increases. Many improvements based on ResNet have been proposed, such as ResNeXt, DenseNet, MobileNet, and ShuffleNet. Adopting the great benefits of Muti-views CNN and mature implementation of ResNet, our framework is capable of offering a remedy for the small samples of source data, which is common in medical imaging field.

\subsection {Depth Representation for 3D Object}

Recent developments in depth imaging sensors have induced an effective application of depth cameras which are actively used for a variety of image recognition tasks~\cite{xu2013exploringhumanactivityrecognition}. The depth camera is used to produce high quality depth images and is getting more significance in multimedia analysis and man-machine interaction~\cite{multipledepthcameras}. The major principle of depth imaging system is extracting a depth target object silhouette and discarding the background, which renders insensitivity to lighting conditions and offers more spatial characteristics. Depth camera is proved to be helpful for many tasks by its benefits. For example, Papazov et al.~\cite{papazov2015real} employ a commodity depth sensor for 3D head pose estimation and facial landmark localization. Kamal et al.~\cite{kamal2016depthHMM} describe a novel method for recognizing human activities from video using depth silhouettes. Raghavendra et al.~\cite{raghavendra2015lightfieldcamera} present a novel face recognition system by focusing on the multiple depth images rendered by the Light Field Camera. Ge et al.~\cite{ge20163d_hand_Pose} generate 2D multi-view projections from a single depth image and fuse to produce a robust 3D hand pose estimation. They have observed the transformation of 3D model, depth image, and multiple 2D projections, which is practicable and flexible. The idea has also been employed in a multi-view video coding system~\cite{merkle2007video}.

Depth representation can also play an important role in assisting detection or surface segmentation tasks combining with RGB images~\cite{deng2017amodal, hoiem2015surface}. Such studies all take advantage of the strong relevancy between the depth representation and 3D model. In other words, these indicate one specific subject. And the flexibility of such transformation can be utilized for improving the algorithm performance.

In our tasks, we exploit our medical imaging data and make use of the depth representation of facial part for human recognition.

\section{Proposed Approach}

\begin{figure*}
  \begin{center}
  \includegraphics[width=1\linewidth]{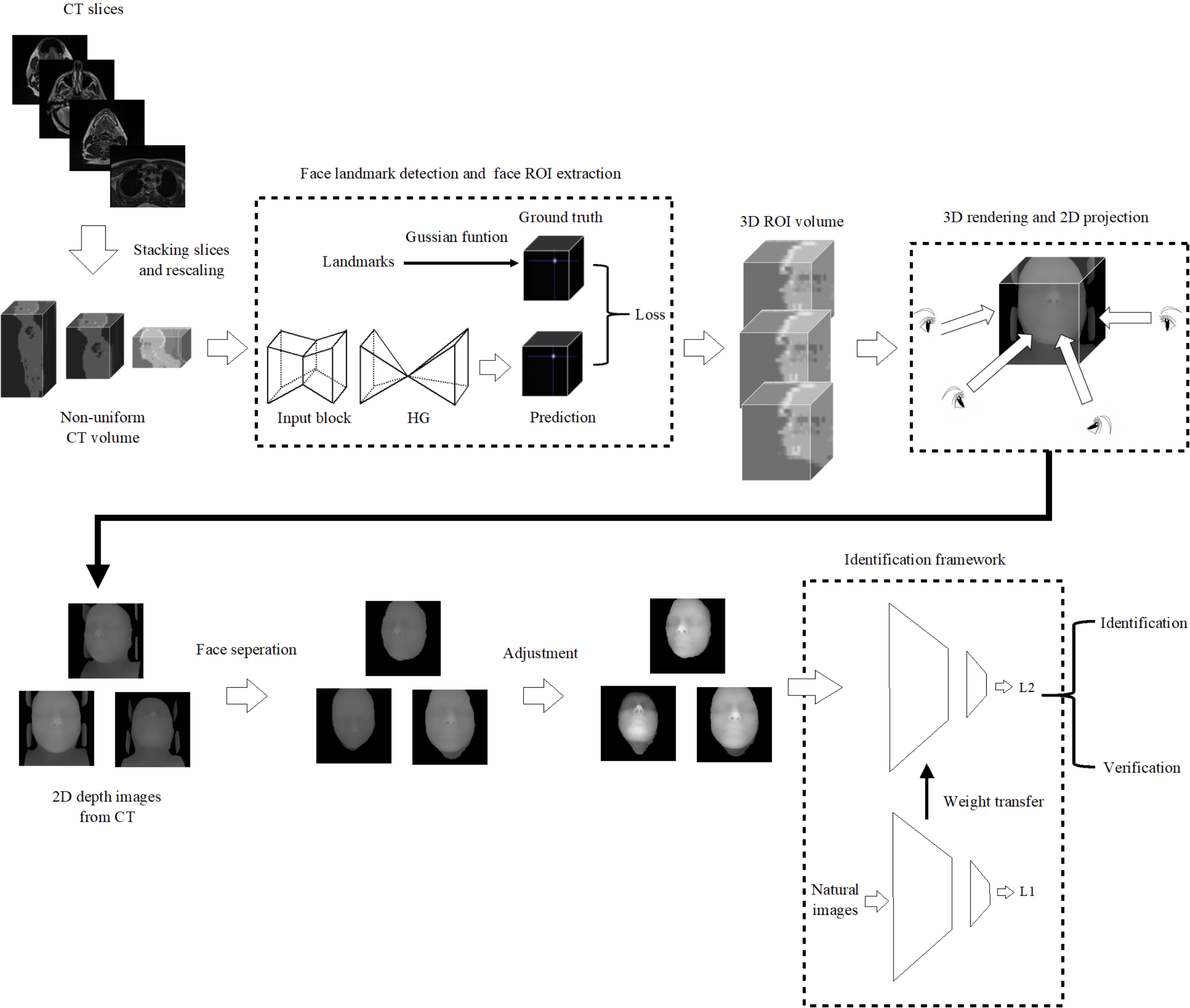}
  \caption{An overview of the proposed approach for 3D CT based face recognition.}
  \label{Total}
  \end{center}
\end{figure*}

In this paper, we propose an effective approach for exploring the biometric features in CT images for face recognition. We first normalize different 3D CT images into the same spacing and organize them in a 3D manner. Then, we detect 3D facial landmarks from 3D CT volumes by learning a modified face alignment network (denoted as FAN below). The detected 3D facial landmarks are then used for cropping the facial region from CTs. Next, 3D rendering and 3D-to-2D projection are performed to obtain the face depth images, which will be used for face recognition. Finally, we leverage transfer learning to adapt a pretrained face recognition model from face depth images in the RGB-D domain to the computed face depth images from CTs. The overview of our approach is illustrated in Fig.~\ref{Total}. Our method can exploit the uniqueness of CT images into consideration and learn the most discriminative feature for face recognition tasks. We adopt transfer learning and a novel training strategy to improve performance. We describe the detailed procedure below.

\subsection {CT Normalization and Face Landmark Detection}
While CTs are widely used in medical diagnosis, there are big diversities among different 3D CT images. First, the slice spacing of 3D CT images captured by different CT imaging machines are usually different, i.e., the slice spacing for CT images can vary from about $1mm$ to $4mm$ in practice. Second, the body areas presented in different CT scans are usually different. For example, some CT scans may contain body from the legs to the head, but some may just contain the head. Therefore, our first step is to extract the facial regions from these various CT scans, and normalize their slice spacing to the same scale.

We extract facial regions from CTs by detecting facial landmarks as widely used in face recognition. However, while facial landmark detection in 2D and 3D face images has been widely studied~\cite{zhu2016face, zhang2014facial}, facial landmark detection from CTs is a relatively new problem because of the modality difference between CTs and traditional 2D and 3D face images.
So, we propose a CT facial alignment network (CT-FAN) to localize a set of pre-defined facial landmarks from each CT image. Our CT-FAN consists of an input block with 4 convolution layers and an Hourglass (HG)~\cite{newell2016stackedhourglass} block. Following the widely used settings in visible RGB image-based face recognition~\cite{valle2019facealignment, Dong2018StyleAN}, we also pre-define facial landmarks on each 3D CT face image, including left eye center, right eye center, nose tip, and left mouth corner. We have manually annotated these landmarks for all the CT images using 3D Slicer~\cite{3D_Slicer}. We denote the landmarks for each CT image as $P=\{p_j\}$, in which $p_j = [x_j, y_j, z_j]$ denoting a coordinate in 3D space. Then, landmark detection from CTs can be formulated as
\begin{equation}\label{CTFAN}
P = \psi_{CT-FAN} (X),
\end{equation}
where $X$ denotes an input CT image. Considering the ambiguity of landmarks in 3D CTs, instead of directly predicting the coordinate values of landmarks, we represent each landmark with a Gaussian blob (or heatmap) centered at the landmark, i.e., $s_j = G(p_j)$. Then, the goal of our CT-FAN becomes
\begin{equation}\label{CTFAN2}
S = \psi_{CT-FAN} (X),
\end{equation}
where $\psi_{CT-FAN}$ can be learned with the conventional Adam solver.
After $\psi_{CT-FAN}$ is learned, given a testing CT image $X$, its predicted extended landmarks can be denoted as $\hat{S}$. Then, the final facial landmarks $\hat{P}$ can be computed with an average of the 3D points in $\hat{S}$. A simple mean square error (MSE) loss is adapted for minimizing the differences.
\begin{equation}\label{MESloss}
L_{CT-FAN} =\frac{1}{2n} \sum{\| P - \hat{P} \|^2}
\end{equation}
Then, the facial ROI volume from each CT image $X'$ can be easily computed based on the predicted landmarks and the bounding box.

\begin{figure}
  \begin{center}
  \includegraphics[width=3.3in]{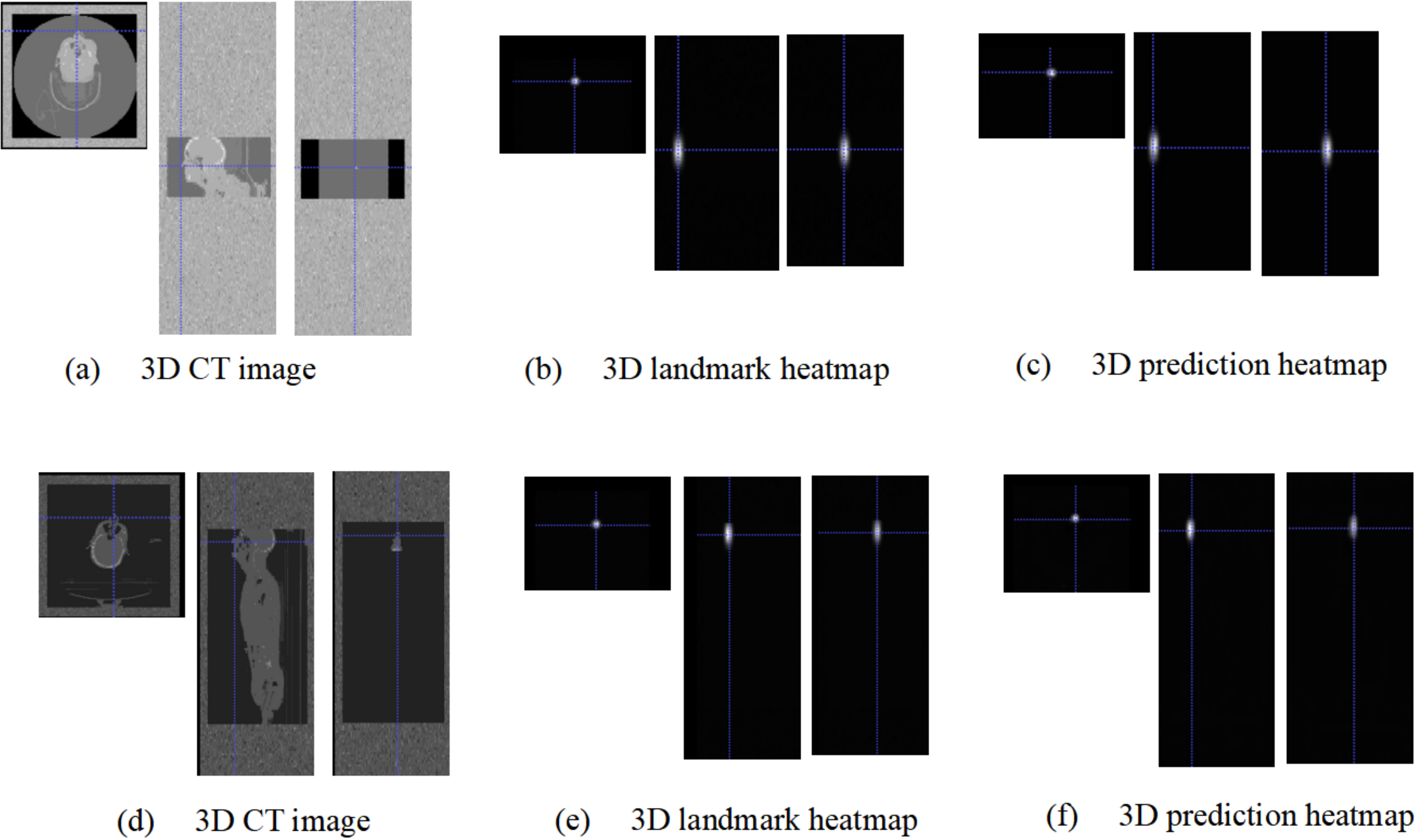}
  \caption{The results of face landmark detection. (a)-(c) are nose landmark detection results from a 3D CT image, and (d)-(f) are nose landmark detection results from another 3D CT image.}
  \label{face_detect}
  \end{center}
\end{figure}

\subsection {3D Rendering and Depth Image Generation}
After obtaining facial ROI volume $X'$, we perform 3D rendering and 3D-to-2D projection to generate CT depth images which will be used for the final face recognition task. Due to the diversity of slice spacing among CTs, scale normalization is crucial. However, the conventional scaling method by utilizing interpolation may lead to voxel holes (i.e., no value assigned to voxels). Therefore, we employ the visualization toolkit (VTK)~\cite{VTK, gpuVTK_applications} tool and construct a 3D rendering model for each CT face image. Compared to other visualization methods for natural images, VTK is more suitable for medical image management and better restoring the original 3D contour information of subjects.

The 3D rendering of a CT face image can be written as:
\begin{equation}\label{3Drender}
CX = \psi_{REN}\{ X' | \gamma, \theta \},
\end{equation}
where $\psi_{REN}$ indicates 3D rendering. $\gamma$ and $\theta$ indicate iso-surface extraction threshold and surface normal rendering angle for 3D model construction, respectively. Specifically, we set $\theta =60^{\circ}$, and $\gamma$ ranges in [-400, -300].

Then, we project the rendered 3D CTs to 2D to obtain CT face depth images. The key of projection is to find the relationship between 2D and 3D coordinates. VTK is capable of simulating the action of taking pictures in reality, like  ``photographing'', and it can transform the coordinates naturally. Assuming $CX$ consists of a set of 3D voxels ${(x_j, y_j, z_j)}$ coordinate. Then, we can obtain the depth image $DX$ with a set of 2D pixels $({x_j}^{'},{y_j}^{'})$ as follows:
\begin{equation}\label{camera2}
DX = \psi_{TS}(CX|cl, co, sc, u_0, v_0, u, v)
\end{equation}
where $\psi_{TS} ()$ is a transformation function, which is defined based on a number of parameters, like camera location $cl$, camera orientation $co$, scaling parameter $sc$, the coordinates of the image center $(u_0, v_0)$ and the size of the 2D images $(u, v)$. Specifically, we set $co$ as default, which means facing the object. We set $u=v=256, u_0=v_0=128$, and $sc=180^{\circ}$.

The parameter setting is largely related to our task. In a common CT scan, there exist different rotation degrees of the human head, e.g., some patients will look up or bow, and others will turn right or left. In addition, our dataset has a limited number of images per patient (most individual only has one CT). Such situation causes difficulties in acquiring CTs of different head pose. Therefore, we utilize different camera locations for data augmentation to simulate different head poses. Camera location $cl$ is determined by three types of rotation, i.e., pitch rotation angle $rp$, roll rotation $rr$, and yaw rotation $ry$.  Specifically, we set $rp\in [-20, 20]$, and $rr\in [-25, 25]$ and $ry=0$. For each 3D CT image, we obtain 90 depth images with different head poses.

\subsection {Face Separation}
In order to suppress the influence of non-facial noises (neck or hands) and get more distinguishable CT images, we then adopt face segmentation. A fully convolutional encoder-decoder network (CEDN)~\cite{CEDN} is employed for face contour extraction. We utilize VGG16~\cite{Simonyan15VGG} as a basic framework and modified it as an encoder. The decoder part is a corresponding deconvolution neural network.

We first train the model on the Helen~\cite{Helen} dataset, which includes 2,000 face images with 68 landmarks, like eyes, nose, mouth, eyebrows, and chin, et al. Then the network is transferred to depth images of CTs.

The separated depth images $SX$ can be obtained as follows:
\begin{equation}\label{2Dprojection}
SX =\psi_{CEDN}(DX)
\end{equation}
where $\psi_{CEDN}$ represents the output of CEDN network. 

\subsection {Image Adjustment}
\begin{figure}
  \begin{center}
  \includegraphics[width=3in]{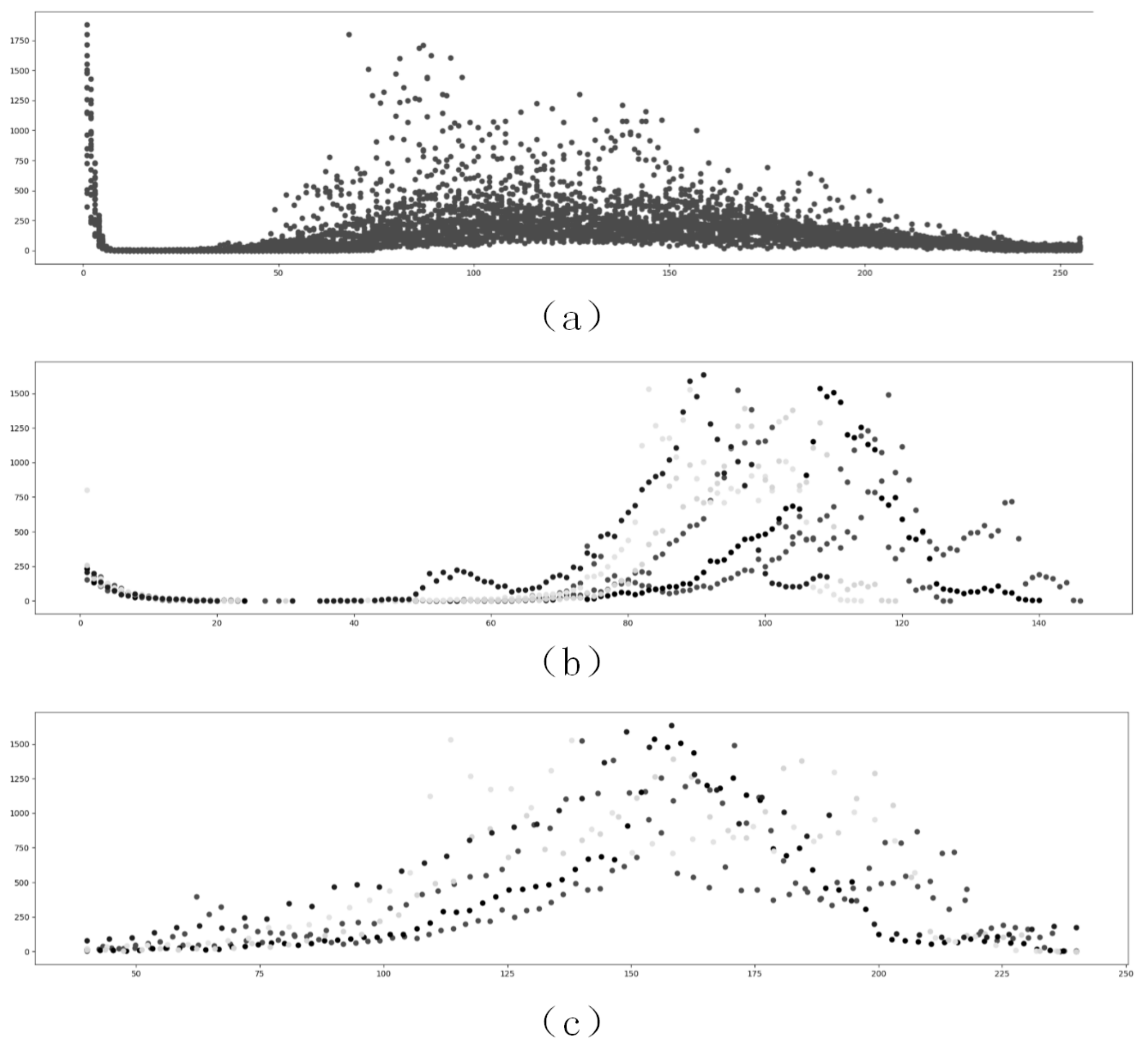}
  \caption{The procedure and results of image adjustment. (a) and (b) are data distribution maps from source domain and target domain, respectively. (c) is the normalization results.}\label{fig_adjust}
  \end{center}
\end{figure}

Due to the limited number of samples, which is common in medical image analysis, training -from-scratch is insufficient to accomplish complex assignments. We employ transfer learning and adopt a depth image dataset collected in natural environment as source dataset for better performance. To reduce the gap between the source and target domain, we analyze the data distribution of the two datasets. The analysis results are shown in Fig.~\ref{fig_adjust}(a) and Fig.~\ref{fig_adjust}(b). It can be discovered that the pixel values of natural depth images are gathered in $40\sim 220$, while the pixel values of the CT depth images are gathered in $80\sim 140$. Then, we select an appropriate threshold and normalize the depth images. The normalization function is described as:
\begin{equation}\label{norm}
NX = C \frac{SX - \theta}{\max(SX) - \theta} 
\end{equation}
where $C$ is a scaling parameter and $\theta$ is threshold. The adjusted result is shown in Fig.~\ref{fig_adjust}(c). The normalized image $NX$ is used as the input to the network.

\subsection {Classification Model and Training Strategy}
We introduce a modified ResNet50~\cite{he2016resnet} as our backbone for face identification and verification tasks, and utilize triplet loss as a loss~\cite{schroff2015facenet} function for limited samples. To this end, the anchor, positive, and negative samples in each batch are required to select appropriately. The loss function for face recognition task is formulated as:
\begin{equation}\label{equa_lossfunc}
L\!=\sum{
abs\begin{pmatrix}\begin{Vmatrix}
X_a\!-X_p
\end{Vmatrix}_2
\!-
\begin{Vmatrix}
X_a\!-X_n
\end{Vmatrix}_2
\end{pmatrix}\!+m},
\end{equation}
where $abs()$ represents the absolute value; $X_a$, $X_p$ and $X_n$ represent anchor, positive sample and negative sample of each batch, respectively, and $m$ is a margin threshold.

To eliminate the error for determining the three samples of loss calculation, the diversity of each batch is largely required. Since our datasets are collected from multiple sources, there is a possible dissimilarity between 3D CTs of different patients. We employ a training strategy for 3D CTs, which forces the network to focus on the inter-individual differences of CTs. We first divide depth images of the same patient into $G_k$ random subgroups, where $k$ is patient ID (denote as label). The size of subgroups $E$ is related to the batch size. Then, $L$ labels are randomly selected. Finally, each batch consists of one random subgroup from each selected label.

\section{Experiments}

\subsection{Dataset and Protocol}
Our CT dataset consists of five open-source head and neck datasets from TCIA Collections\footnote{https:/www.cancerimagingarchive.net}, including Head-Neck-PET-CT, QIN-HEADNECK, HNSCC-3DCT-RT, CPTAC-HNSCC, and Head-Neck Cetuximab. The Cancer Imaging Archive (TCIA) is a large archive of medical images. Some datasets also offer other image-related information such as patient treatment results, treatment details, genomics and pathology related information, and expert analysis.

Head-Neck-PET-CT~\cite{head-and-neck} consists of 298 patients from FDG-PET/CT and radiation therapy programs for head and neck (H\&N) cancer patients supported by 4 different institutions in Quebec. All patients underwent FDG-PET/CT scans between April 2006 and November 2014. The same transformation is applied to all images, preserving the time interval between serial scans. The patient treatments and image scanning protocols are elaborated in \cite{head-and-neck}. QIN-HEADNECK is a set of multiple positron emission tomography/computed tomography (PET/CT). 18F-FDG scans-before and after therapy-with follow up scans of head and neck cancer patients. The data contributes to the research activities of the National Cancer Institute's (NCI's) Quantitative Imaging Network (QIN). HNSCC-3DCT-RT consists of 3D high-resolution fan-beam CT scans of 31 head-and-neck squamous cell carcinoma (HNSCC) patients by using a Siemens 16-slice CT scanner with standard clinical protocol. CPTAC-HNSCC contains subjects from the National Cancer Institute’s Clinical Proteomic Tumor Analysis Consortium Head-and-Neck cancer (CPTAC-HNSCC) cohort. Head-Neck Cetuximab is a subset of RTOG 0522/ACRIN 4500. The protocol is randomized chemotherapy and phase III trial of radiation therapy for stage III and IV head and neck cancer.

In total, 600 3D CT images of 280 patients are collected after integrating and screening the above five datasets. We randomly select 224 subjects for training and the remaining 56 subjects for testing. We obtain a total of 54,000 CT depth images through data augmentation. Since the number of CT depth images of different classes may be different, we divide the training and testing sets with a similar ratio for each class, i.e., split each class by 8:2.

\begin{figure}
  \begin{center}
  \includegraphics[width=3.0in]{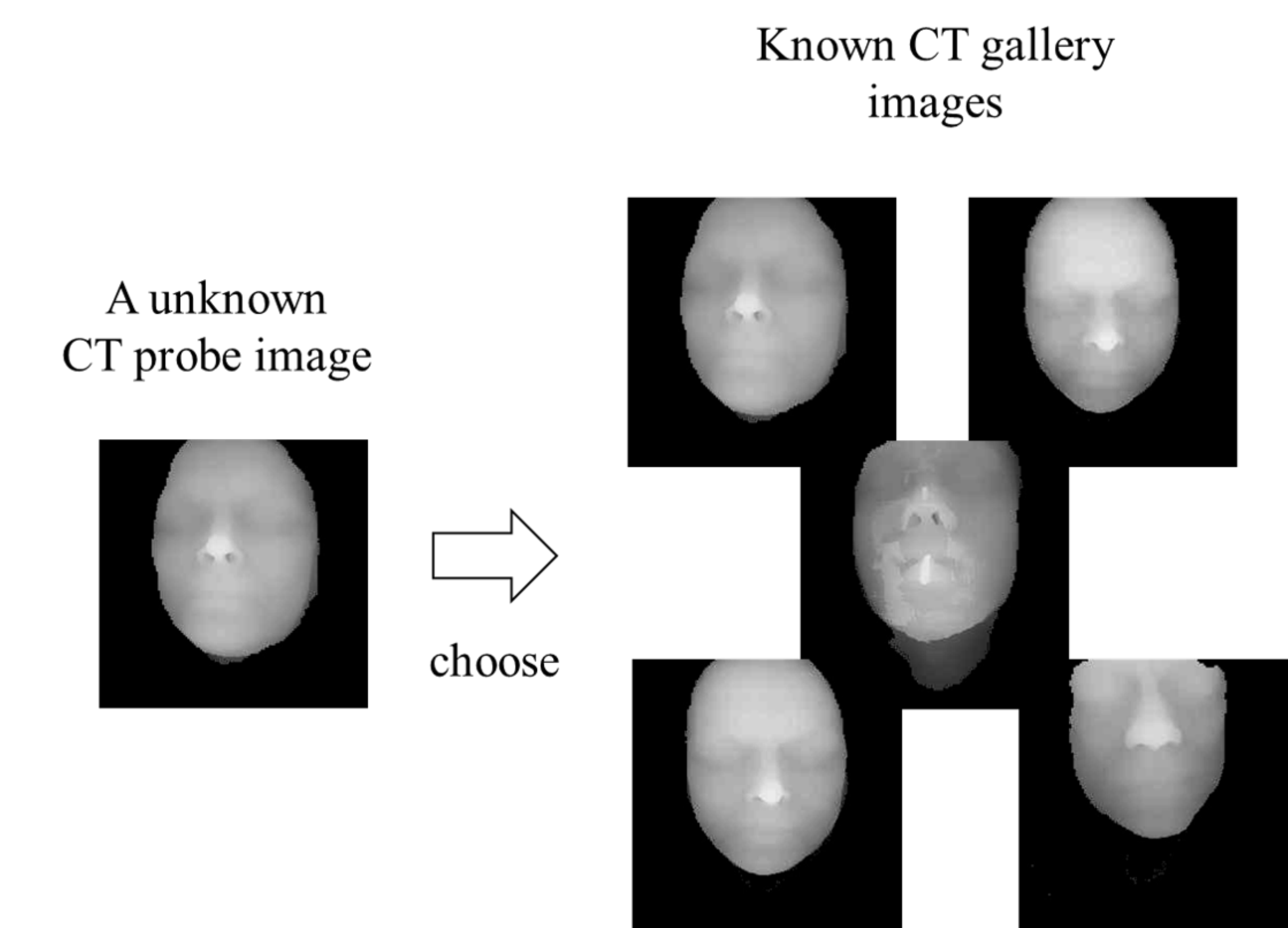}\\
  \caption{An example of a CT probe face image and several gallery CT face images after our data processing pipeline.}
  \label{fig_gallery_probe}
  \end{center}
\end{figure}

We perform both face identification and verification experiments using the CT depth images obtained by our approach to verify the effectiveness of leveraging CT depth images for human identification. Firstly, we perform face identification and calculate the classification or identification accuracy (ACC) for performance evaluation. One sample from each class in test dataset is selected to the gallery, and the remaining CT depth images are used as probes. We measure the distance between each probe and all gallery samples. And class of the gallery with the smallest distance is regarded as predicted label. Our gallery and probe sets contain 54 CT depth images and $700\sim 800$ CT depth images, respectively. Face identification is a simulation of 1 vs. N matching tasks for an unknown 3D CT in reality. The illustration is shown in Fig.~\ref{fig_gallery_probe}. In addition, we perform face verification to replicate the 1 vs. 1 face matching scenarios. Specifically, given a random pair of CT depth images, it is called genuine pair if the two images are from the same subject; otherwise, the pair is called impostor pair. The verification accuracy (VACC) and the area under the ROC curve (AUC) is adopted for evaluation. The threshold calculated by ROC is utilized as indicator for positive and negative classification.

\begin{table*}[!tb]
\caption{Face identification and verification results of our method and the other baseline methods. `Pretrain' indicates if a pretrained model is used or not. Mean ACC, Mean VACC and Mean AUC indicates the mean identification accuracy, the mean verification accuracy and the mean area under ROC curve, respectively. The standard deviations of the 5-fold test are shown in brackets.}
\label{table_result}
\centering
\begin{tabular}{cccccccc}
\hline
Model                              & Cropping & Rotation & Pretrain & Mean ACC (std)     & Mean VACC (std)    & Mean AUC (std)     & Data format     \\
\hline\hline
VGG16                              &          &          & n        & .2482($\pm$0.026) & .6733($\pm$0.028) & .7518($\pm$0.023) & 3D volume       \\
VGG16                              &          & $\surd$  & n        & .2024($\pm$0.059) & .6693($\pm$0.021) & .7272($\pm$0.017) & 3D volume       \\
VGG16                              & $\surd$  &          & n        & .2461($\pm$0.085) & .6951($\pm$0.047) & .7589($\pm$0.052) & 3D volume       \\
VGG16                              & $\surd$  & $\surd$  & n        & .3444($\pm$0.041) & .7568($\pm$0.033) & .8265($\pm$0.022) & 3D volume       \\
ResNet10                           &          &          & n        & .2930($\pm$0.068) & .6612($\pm$0.034) & .7301($\pm$0.018) & 3D volume       \\
ResNet10                           &          & $\surd$  & n        & .2820($\pm$0.044) & .6829($\pm$0.030) & .7446($\pm$0.039) & 3D volume       \\
ResNet10                           & $\surd$  &          & n        & .2863($\pm$0.060) & .7210($\pm$0.009) & .7989($\pm$0.021) & 3D volume       \\
ResNet10                           & $\surd$  & $\surd$  & n        & .3571($\pm$0.085) & .7608($\pm$0.026) & .8423($\pm$0.029) & 3D volume       \\
ResNet18                           &          &          & n        & .3236($\pm$0.078) & .6831($\pm$0.027) & .7597($\pm$0.019) & 3D volume       \\
ResNet18                           &          & $\surd$  & n        & .2837($\pm$0.048) & .6747($\pm$0.051) & .7342($\pm$0.04)  & 3D volume       \\
ResNet18                           & $\surd$  &          & n        & .2877($\pm$0.045) & .7079($\pm$0.032) & .7883($\pm$0.031) & 3D volume       \\
ResNet18                           & $\surd$  & $\surd$  & n        & .2775($\pm$0.076) & .7318($\pm$0.043) & .8052($\pm$0.041) & 3D volume       \\
ResNet34                           &          &          & n        & .3407($\pm$0.049) & .7234($\pm$0.029) & .7893($\pm$0.038) & 3D volume       \\
ResNet34                           &          & $\surd$  & n        & .2930($\pm$0.042) & .6831($\pm$0.032) & .7379($\pm$0.027) & 3D volume       \\
ResNet34                           & $\surd$  &          & n        & .3271($\pm$0.097) & .7620($\pm$0.024) & .8032($\pm$0.058) & 3D volume       \\
ResNet34                           & $\surd$  & $\surd$  & n        & .4000($\pm$0.094) & .7771($\pm$0.056) & .8535($\pm$0.048) & 3D volume       \\
PointNet                           & -        & -        & n        & .3634($\pm$0.054) & .7089($\pm$0.039) & .7263($\pm$0.038) & Point cloud     \\
PointNet (ModelNet40 pretrained)   & -        & -        & y        & .3762($\pm$0.045) & .6786($\pm$0.010) & .7529($\pm$0.097) & Point cloud     \\
PointNet++                         & -        & -        & n        & .2066($\pm$0.013) & .6308($\pm$0.009) & .6907($\pm$0.014) & Point cloud     \\
PointNet++ (ModelNet40 pretrained) & -        & -        & y        &          .2055($\pm$0.017)  & .6355($\pm$0.009) & .6965($\pm$0.13)  & Point cloud     \\
\hline
Proposed                           & -        & -        & n        & .8712($\pm$0.032) & .9374($\pm$0.009) & .9857($\pm$0.004) & CT Depth images \\
Proposed (ImageNet pretrained)     & -        & -        & y        & .8764($\pm$0.051) & .9513($\pm$0.027) & .9849($\pm$0.005) & CT Depth images \\
Proposed (Depth images pretrained) & -        & -        & y        & \textbf{.9253($\pm$0.044)} & \textbf{.9612($\pm$0.016)} & \textbf{.9936($\pm$0.004)} & CT Depth images \\
\hline
\end{tabular}
\end{table*}

\begin{figure*}[!t]
  \begin{center}
  \includegraphics[width=\linewidth]{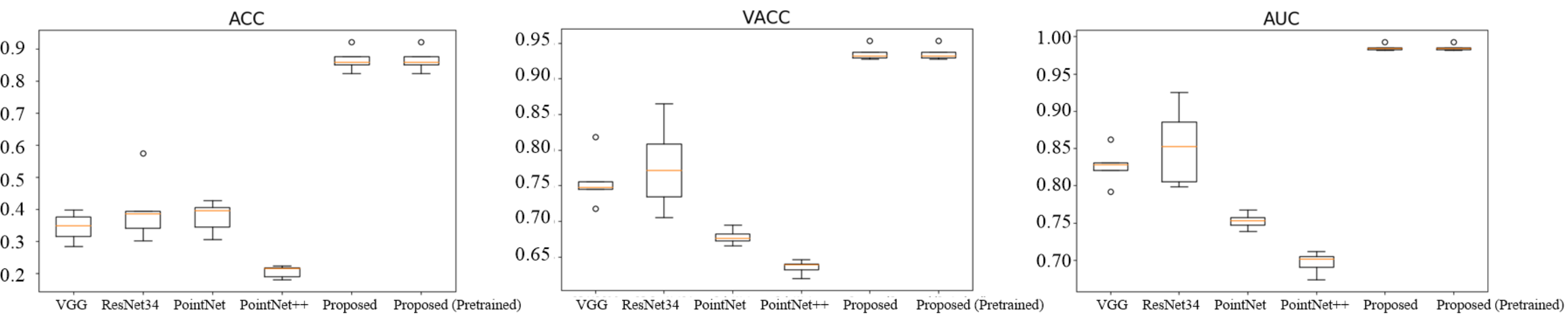}\\
  \caption{The box plot figures of face identification and verification. Four baselines, e.g., VGG, ResNet34, pretrained Pointnet, pretrained PointNet++ and our proposed model are shown.}
  \label{box_plot}
  \end{center}
\end{figure*}

\begin{figure*}[!t]
  \begin{center}
  \includegraphics[width=5.5in]{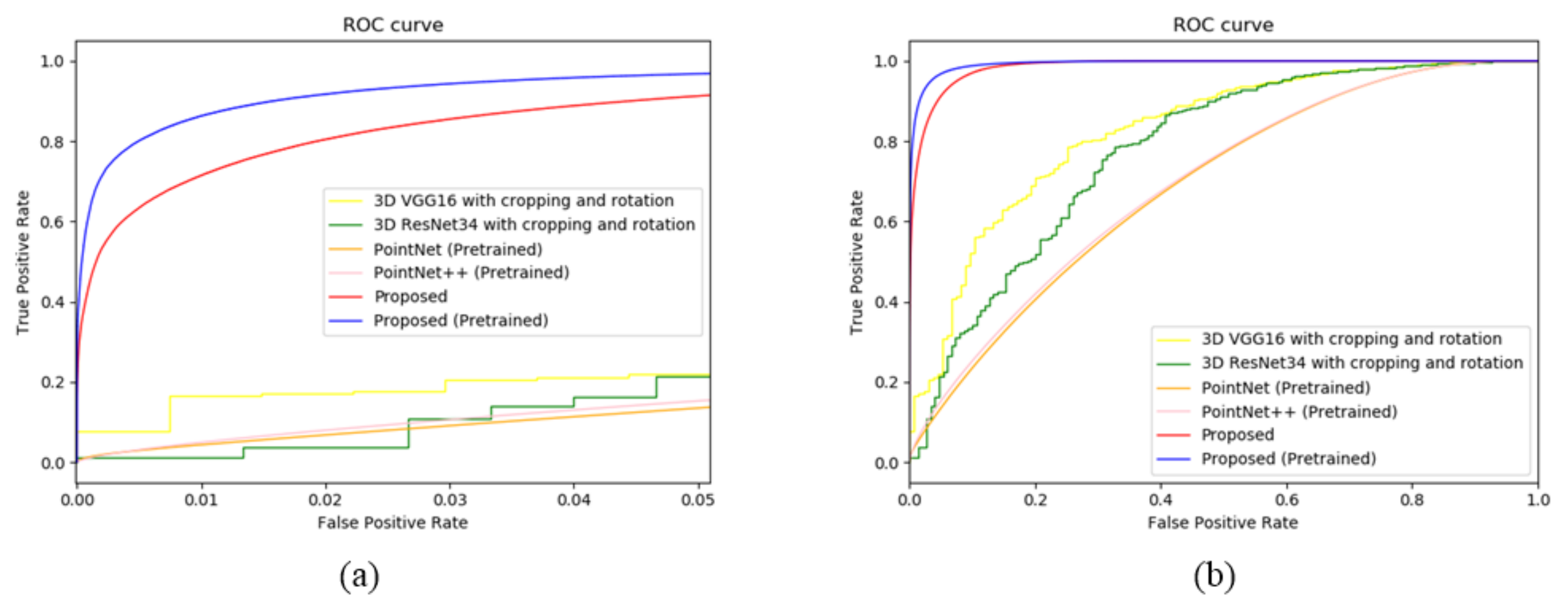}\\
  \caption{The ROC curves of face verification, in which (a) highlights the true positive rates (TPR) at false positive rates (FPR) ranging in [0, 0.05], and (b) shows the whole ROC for FPR ranging in [0, 1].}
  \label{ROC}
  \end{center}
\end{figure*}

\begin{table*}[!tb]
\caption{Face identification and verification results of our method with different training settings. Grouping and proportion are related to different selections of training strategy. Grouping setting, Pretrain and Data stages indicate different parameters (E and L), a pretrained model is used or not, and different data processing stages, respectively.}
\label{table_result2}
\centering
\begin{tabular}{ccccccccc}
\hline
Row ID & Grouping                 & Proportion               & Grouping setting (E*L) & Pretrain           & Data stages            & Mean ACC (std)     & Mean VACC (std)    & Mean AUC (std)     \\ \hline\hline
1& \multirow{2}{*}{$\surd$} & \multirow{2}{*}{$\surd$} & \multirow{2}{*}{15*18} & \multirow{2}{*}{y} & Projected              & .8778($\pm$0.034) & .9419($\pm$0.024) & .9708($\pm$0.020) \\
2&                         &                          &                        &                    & Segmented              & .8842($\pm$0.048) & .9543($\pm$0.024) & .9831($\pm$0.009) \\ \hline\hline
3& \multirow{6}{*}{$\surd$} & \multirow{6}{*}{$\surd$} & 3*90                   & \multirow{6}{*}{y} & \multirow{6}{*}{Final} & .8512($\pm$0.035)    & .9312($\pm$0.012)  & .9785 ($\pm$0.008)   \\
4&                      &                          & 5*54                   &                    &                        & .8517($\pm$0.024) & .9745($\pm$0.015)  & .9208($\pm$0.008)  \\
5&                       &                          & 10*27                  &                    &                        & .8928($\pm$0.030)
                         & .9349($\pm$0.013)  & .9807($\pm$0.002) \\
6&                     &                          & 18*15                  &                    &                        & .8652($\pm$0.032) & .9316($\pm$0.008)  & .9819($\pm$0.004)     \\
7&                       &                          & 30*9                   &                    &                        & .8709($\pm$0.027) & .9417($\pm$0.007)   & .9821($\pm$0.009)  \\
8&                       &                          & 90*3                   &                    &                        & .8648($\pm$0.031) & .9267($\pm$0.010) & .9797($\pm$0.006) \\ \hline\hline
9&                       &                          & \multirow{6}{*}{15*18} & n                  & \multirow{6}{*}{Final} & .7485($\pm$0.025) & .8956($\pm$0.009) & .9632($\pm$0.009) \\
10&                       &                          &                        & y                  &                        & .8670($\pm$0.020) & .9402($\pm$0.009) & .9860($\pm$0.005) \\
11& $\surd$                  &                          &                        & n                  &                        & .8192($\pm$0.032) & .9170($\pm$0.016) & .9771($\pm$0.006) \\
12& $\surd$                  &                          &                        & y                  &                        & .9097($\pm$0.012) & .9471($\pm$0.015) & .9888($\pm$0.006) \\
13&                     & $\surd$                  &                        & n                  &                        & .8377($\pm$0.034) & .9328($\pm$0.012) & .9809($\pm$0.003) \\
14&                      & $\surd$                  &                        & y                  &                        & .8516($\pm$0.031) & .9325($\pm$0.011) & .9825($\pm$0.004) \\ \hline\hline
15& \multirow{2}{*}{$\surd$} & \multirow{2}{*}{$\surd$} & \multirow{2}{*}{15*18} & n                  & \multirow{2}{*}{Final} & .8712($\pm$0.032) & .9374($\pm$0.009) & .9857($\pm$0.004)  \\
16& &                          &                        & y                  &                        &  \textbf{.9253($\pm$0.044}) &\textbf{.9612($\pm$0.016)} & \textbf{.9936($\pm$0.004)}\\\hline
\end{tabular}
\end{table*}

\begin{figure}[!tb]
  \begin{center}
  \includegraphics[width=3in]{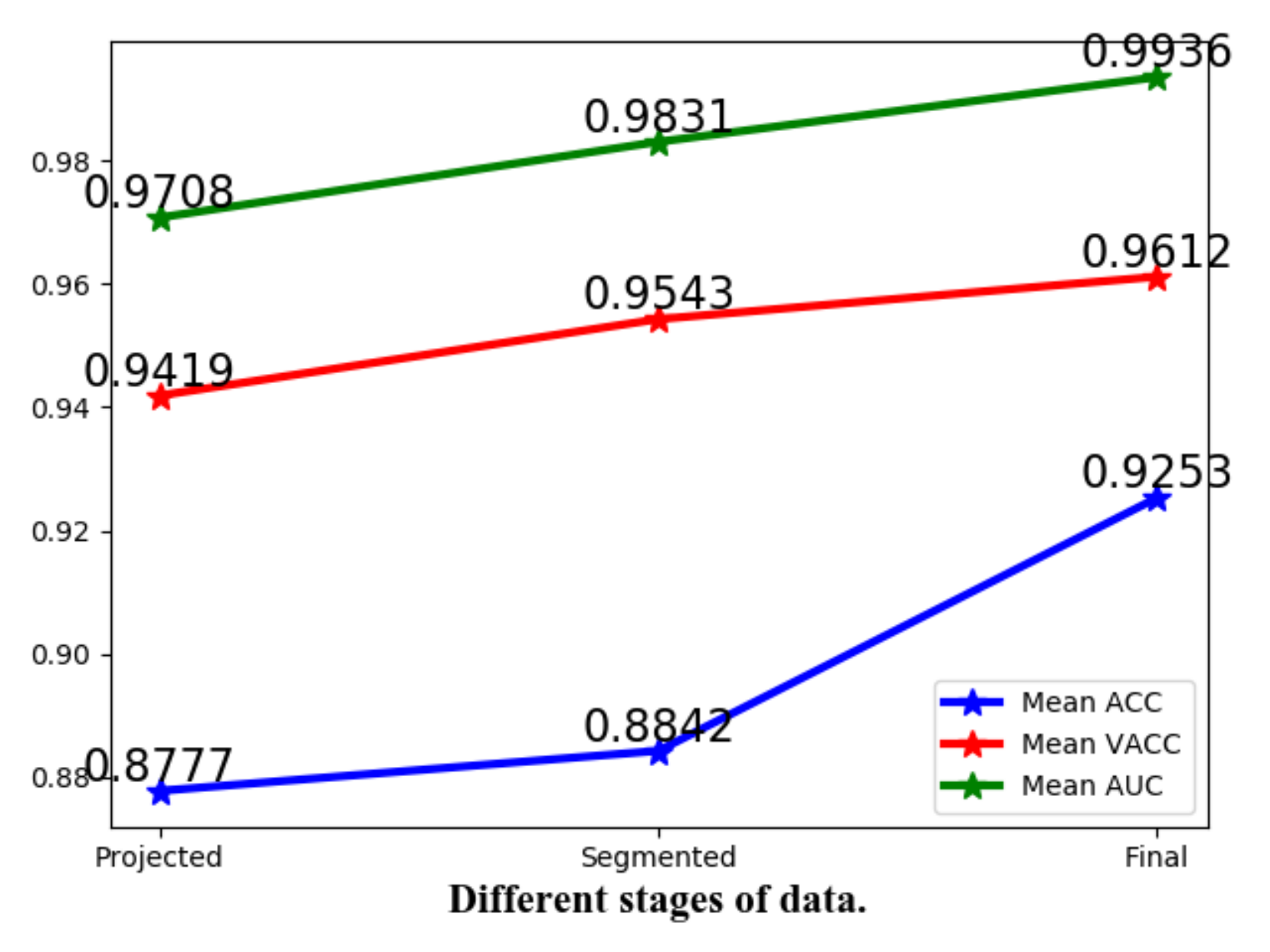}\\
  \caption{Face identification and verification results on our CT dataset when gradually adding the individual steps of our processing pipeline.}
  \label{results_data_stages}
  \end{center}
\end{figure}

\begin{figure*}
  \begin{center}
  \includegraphics[width=6in]{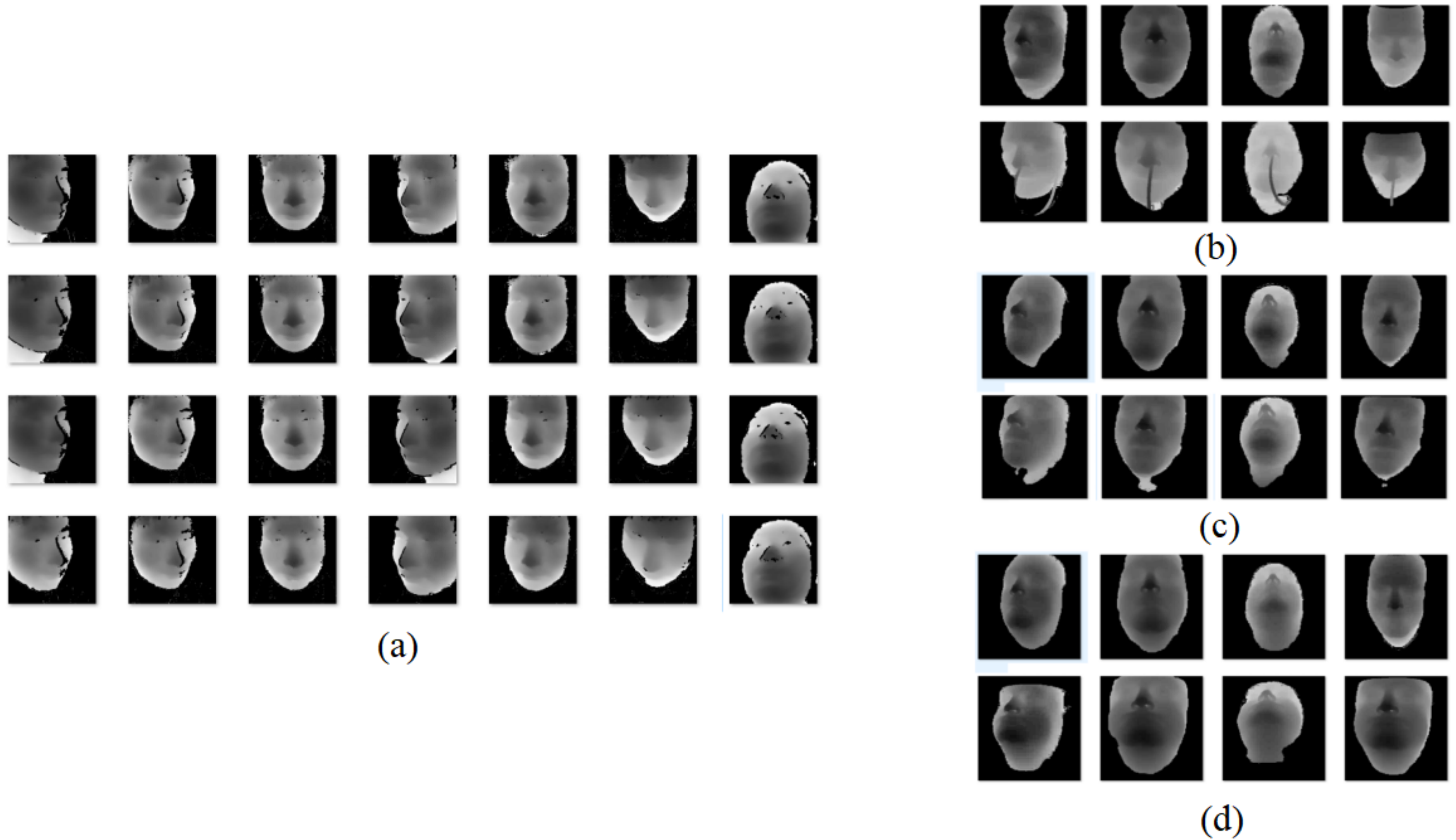}\\
  \caption{Examples of (a) are the source domain face depth images of one subject from the RGB-D dataset in \cite{Cui_FG}, and (b-d) are the 2D face depth images generated from 3D CTs for three subjects.}\label{dis}
  \end{center}
\end{figure*}

\subsection{Implement Details}
We employ transfer learning to mitigate the challenges caused by small sample of our CT dataset. We first train our classification model ResNet50 by using depth images from an RGB-D face dataset, which contains 581,366 depth images of different shooting angles from 450 subjects~\cite{Zhang_RGBD, Cui_FG}. Then, the model is fine-tuned to CT depth images. 

We set $E=15$ and $L=18$ for training strategy, and use a batch size of $270$ in our experiments. The CT depth images are normalized to $256\times 256$ with a $224\times 224$ random crop as input. The classification model for CT depth is implemented using PyTorch. The initial learning rate is set to $1e^{-3}$. The Adam solver is used as the optimizer for network training.

For the baseline methods of PointNet and PointNet++, we extract point cloud from CT depth images so that they can be input to the baseline methods. Specifically, we downsample 4096 points from each depth image as input, and set the batch size to 48. For 3D volume-based baseline methods like 3D VGG, 3D ResNet10, 3D ResNet18, and 3D ResNet34, the 3D CTs of different pre-processing methods are fed into 3D network directly. We utilize multi-scale grouping (MSG) model for PoinNet++. We employ 5-fold cross validation for all baseline methods and our model.

\subsection{Results of Our Model and Baselines.}
Firstly, we compare our method with other 3D based methods for 1 vs. 54 identification task, such as PointNet, PointNet++ and 3D ResNet. 
The results are summarized in Table~\ref{table_result}. It is easy to discover that our method can obtain much better performance than the baseline methods.
Specifically, our pre-trained method obtains an identification accuracy of \textbf{92.53\%}, more than \textbf{52.53\%} higher than the best of the baseline method, e.g., 3D ResNet34. And our un-pretrained obtains an identification accuracy of \textbf{87.12\%}, \textbf{50.78\%} higher than the best of the baseline method, e.g., PointNet.
The 3D CNNs are observed to get lower accuracy, which may because of the small number of training samples, since each 3D CT only has one data point. We also evaluate the effectiveness of data augmentation by random 3D rotation for 3D CNN. However, such augmentation proved to be unprofitable, which is affected by 3D deformation. In addition, it is difficult to find a large 3D dataset which is similar to 3D CTs for pre-training. 
In addition, the performances of PointNet and PointNet++ are weak. Because their limited input representation may cause detailed information loss. We also evaluate the pre-trained performances of PointNet and PointNet++ as mentioned in~\cite{qi2017pointnet++}. The transfer results from ModelNet40 are proved to be helpless, which largely due to the gap between CT depth images and 3D natural point cloud. And such point-cloud-based models are also limited in terms of input size. Thus, our model is capable of learning discriminatice feature by its design.
We also draw the box plot and ROC in Fig.~\ref{box_plot} and Fig.~\ref{ROC} by using the best model of each network, e.g., VGG with cropping and rotation, ResNet34 with cropping and rotation, PointNet with ModelNet40 pretrained and PointNet++ with ModelNet40 pretrained. The ROC results suggest that our model leads to the best performance, which shows a robustness feature learning capacity.

We evaluate the performance of 1 vs. 1 verification task. It can be discovered that our method can obtain the best verification performance than other baseline methods. The results are shown in Table~\ref{table_result}. Specifically, our method obtains a verification rate of \textbf{96.12\%}, with \textbf{18.41\%} higher than the best of the baseline method, e.g., ResNet34. Also, the AUC (0.9936 by our proposed method) suggests that our method can yield a discriminative performance, indicating powerful identification abilities for 3D CTs. The 3D volume-based baseline methods and point-cloud-based methods also proved to less beneficial, compared to our model.

\subsection{Results of Different Training Settings.}
To validate the robustness of our training strategy, we also compare our training strategy with other training settings. There are two important operations in our proposed method. First, we shuffle the samples in subgroup generation, and each subgroup could contain depth images generated from different 3D CTs of the same patient. Second, we split the training and testing set according to the number of 3D CTs obtained from the same patient, and we expect to find the superiority and effectiveness of such a partition. To this end, we perform experiments without at least one of the above operations, where other parameters are kept the same to those of the original setting. And results are shown in rows $9\sim 14$ of Table~\ref{table_result2}. We discover that the two operations of our training strategy are beneficial and each operation shows some boost. And the model with two operations can improve accuracy by 3.35\% and 1.56\% over un-pretrained and pretrained models, respectively. Such results indicate that a training method with a suitable partition for training and testing for 3D CTs of multiple sources is crucial.

There are two parameters in our training. i.e., subgroup size $E$ and selected label number $L$. To validate the robustness of our training strategy, we perform different parameter settings with the same batch size $270$. The results are shown in rows $3\sim 8$ of Table~\ref{table_result2}. It can be discovered that the best result is obtained by setting $E=15$ and $L=18$. The triple loss is sensitive to the positive and negative samples. Thus, our design for each mini-batch enables the network to focus on the inter-individual differences of 3D CTs as designed.

In order to evaluate the necessity of our data processing pipeline, we utilize the data generated by each step as network input, including original projected 2D images (Projected), face segmented images (Segmented), and final adjusted images (Final). The results are shown in rows $1\sim 2$ of Table~\ref{table_result2} and Fig.~\ref{results_data_stages}. It can be seen that the final processed data can achieve the best accuracy and AUC value. Each stage can improve the identification performance, e.g. 0.64\% and 4.11\% boost, respectively. The results demonstrate that each step of our model can be essential and can improve the accuracy accordingly.

\section{Discussions}
Transfer learning is employed in our tasks. However, there exists a gap between natural depth images and CT depth images. The samples of natural depth images come from every frame of four videos. And the samples are shown in Fig.~\ref{dis}(a). Each row indicates a video. We discover that the images from different video images have high similarity. It demonstrates the ROI (face area) of a subject is consistent across the different videos.

Meanwhile, we analyze our depth CT images, especially the images generated from different 3D CTs with the same patient. Different from natural depth images, the images are not consistent. For example, the patient may have a tube in his nose when undergoing one CT scans and not have a tube when undergoing the other. The samples are shown in Fig~\ref{dis}(b). Such condition is unusual but exists in the medical field. In addition, the time interval of undergoing multiple CT scans may vary, which can lead to changes of facial shape. The patients underwent CT scans are generally diseased. And they may become more and more emaciated in case of serious conditions. For example, the images in Fig.~\ref{dis}(c) come from the same patient, and different rows indicate different 3D CTs. It is obvious that the images in the first row are thinner than the second row, which leads to a variation of face. The images in Fig.~\ref{dis}(d) come from the same patient as well. We found that the images in the second row have a bigger cheek, and the patient may take a deep breath when scanning. Such conditions may not appear in the laboratory environment. However, our datasets contain the above 3D CTs, which may common in other medical images.

It is difficult for a network to recognize the differences and similarities between an unusual image and usual images. For example, the network may not be able to distinguish the CT depth images with a tube during testing, due to such images not appearing in the training dataset. Our model has its limitation to handle such 3D CTs. Utilizing style transfer learning method to generate more images with different cases may be helpful. The style transfer learning methods largely based on generative adversarial networks (GANs). However, training such networks from 3D CTs may be limited by the small sample problem. And the generated images may not close to the real images.

Our paper employs model transferring from natural depth images. Minimizing the gap between the source domain (natural depth images) and the target domain (CT depth images) is crucial.

\section{Conclusions}
In this paper, we explore the biometric characteristic of 3D CTs and use them to perform face recognition and verification. We propose an automatic processing pipeline for human recognition based on 3D CT. The pipeline first detects facial landmarks for ROI extraction and then project 3D CT faces to 2D to obtain depth images. To address small training data issue and improve the inter-class separability, we use transfer learning and a group sampling strategy to train our classification network from face depth images obtained from 3D CTs. We perform experiments on 3D CTs collected from multiple sources, which shows the proposed method performs better than the state-of-the-art methods like point cloud networks and 3D CNNs. We also validate and discuss the effectiveness of the data processing and transfer learning modules.

Our work is of vital importance to the integrity of medical image big data.  because medical images are inevitably expanding, to assure correct associations between different scans of the same patient it is necessary to leverage automatic human recognition technologies like the proposed CT based face recognition. In addition, the proposed CT based face recognition method can be used to identify unknown CT images when the meta data is lost. Our work can improve the efficiency of hospital operations to a certain extent, avoid cumbersome manual inspections, and bring convenience to hospital visits and medical treatment. Our future work includes combining the CT images with other modalities of the same individual to perform multi-modality or cross-modality human recognition.


\ifCLASSOPTIONcaptionsoff
  \newpage
\fi


\bibliographystyle{IEEEtran}




\vfill
\end{document}